\documentclass[sn-mathphys-num,pdflatex]{sn-jnl}%

\usepackage{url}

\usepackage{graphicx}%
\usepackage{multirow}%
\usepackage{amsmath,amssymb,amsfonts}%
\usepackage{amsthm}%
\usepackage{mathrsfs}%
\usepackage[title]{appendix}%
\usepackage{xcolor}%
\usepackage{textcomp}%
\usepackage{manyfoot}%
\usepackage{booktabs}%
\usepackage{algorithm}%
\usepackage{algorithmicx}%
\usepackage{algpseudocode}%
\usepackage{listings}%

\begin{document}

\title[Article Title]{Intelligent Vacuum Thermoforming Process}

\author[1,2]{\fnm{Andi} \sur{Kuswoyo}}\email{ak2467@cam.ac.uk}
\author[1]{\fnm{Christos} \sur{Margadji}}\email{cm2161@cam.ac.uk}
\author*[1]{\fnm{Sebastian W.} \sur{Pattinson}}\email{swp29@cam.ac.uk}

\affil*[1]{\orgdiv{Department of Engineering}, \orgname{University of Cambridge}, \orgaddress{\street{Trumpington Street}, \city{Cambridge}, \postcode{CB2 1PZ}, \country{UK}}}
\affil[2]{\orgdiv{Faculty of Mechanical and Aerospace Engineering}, \orgname{Institut Teknologi Bandung}, \orgaddress{\street{Jl. Ganesha 10}, \city{Bandung}, \postcode{40132}, \state{Jawa Barat}, \country{Indonesia}}}

\abstract{Ensuring consistent quality in vacuum thermoforming presents challenges due to variations in material properties and tooling configurations. This research introduces a vision-based quality control system to predict and optimise process parameters, thereby enhancing part quality with minimal data requirements. A comprehensive dataset was developed using visual data from vacuum-formed samples subjected to various process parameters, supplemented by image augmentation techniques to improve model training. A k-Nearest Neighbour algorithm was subsequently employed to identify adjustments needed in process parameters by mapping low-quality parts to their high-quality counterparts. The model exhibited strong performance in adjusting heating power, heating time, and vacuum time to reduce defects and improve production efficiency.}


\maketitle

\section{Introduction}\label{sec1}
Vacuum forming is the process of reshaping materials, usually in sheet form, using heat and vacuums. Vacuum thermoforming is one of the oldest and most common methods for thermoplastic processing. This method has long been a staple in the manufacturing industry for efficiently producing extensive thermoplastic components with intricate geometries for diverse applications \cite{Book_SE2012, Hong2022, Correia2022, Valavan2023}. However, despite its widespread use, maintaining consistent quality in vacuum-formed products remains a significant challenge. Ensuring optimal settings is essential to achieving desired product characteristics and mitigating defects such as webbing or surface imperfection. Variations in material properties and tooling configurations can further lead to defects such as sheet thinning or uneven thickness, compromising the integrity and appearance of the final product \cite{Book_PS2012}. Traditionally, quality control in vacuum forming relies on manual inspections and experience-based adjustments, but this can lead to inconsistencies and errors in defect detection, and it can be difficult to inspect all products when large volumes are involved. Hence, there has been growing interest in leveraging technologies such as machine learning to enhance the quality and reliability of vacuum thermoforming processes. Quality assurance and process optimisation can require delicate measurement tools and systems, which can be cost-prohibitive and complex to implement \cite{Rai2021, Ismail2022}. However, vision-based machine learning offers a promising alternative in that it can be information-rich, relatively inexpensive, and easy to set up\cite{Ren2022}. By deploying machine learning algorithms trained on even small image datasets, models can be developed to recognise patterns, features and anomalies indicative of potential defects \cite{LeX2020}. These models can be integrated into manufacturing processes to provide real-time feedback and control, opening the door to reducing defects and optimising product quality. Further development on vision-based machine learning methods can also facilitate predictive maintenance by identifying early warning signs of any degradation, allowing manufacturers to schedule maintenance activities preemptively and avoid costly downtime \cite{Haq2023, Serin2020}. Several studies have explored the application of different algorithms to optimise process parameters, detect defects, and predict product quality for various manufacturing processes \cite{Rai2021, Weichert2019, Kumar2023, Michiels2022, Banus2021}. An important limiting factor in the application of machine learning in manufacturing is often the availability of data. Many models require more data to become performant than can feasibly be produced by available experimental setups. 

\begin{figure}[h]
\centering
\includegraphics[width=1\textwidth]{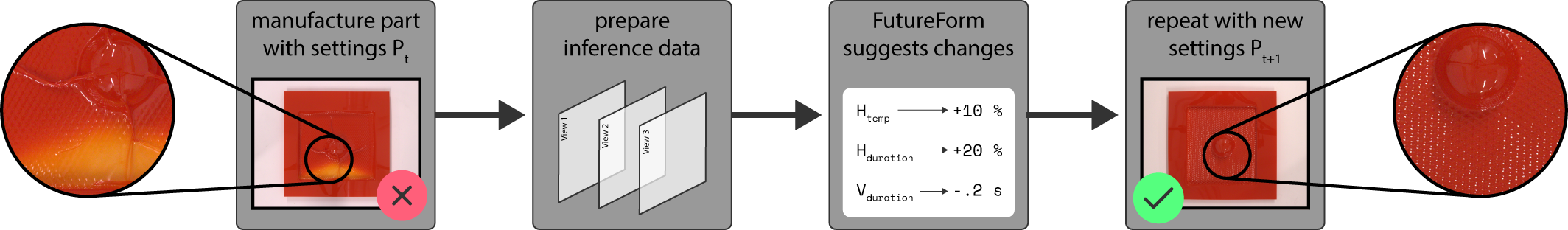}
\caption{Intelligent Vacuum Forming Inference Pipeline}\label{fig1}
\end{figure}

In this work, we integrate k-Nearest Neighbout (k-NN) and Convolutional Neural Networks (CNNs) to develop a robust, vision-based quality control system for vacuum forming that performs well while using relatively little data. Fig.~\ref{fig1} shows the inference pipeline for the proposed method, which begins with producing a part using current parameters. Images are captured and analysed through the model, which then suggests necessary changes to heating temperature, duration, and vacuum time, which are the primary parameters an operator has access to in the system. These adjustments are then applied in subsequent manufacturing cycles to improve part quality. By using this approach, the system can be used for real-time feedback and automated adjustments, enhancing the quality and reducing scrap in vacuum forming.

As with many manufacturing processes, in vacuum thermoforming the optimisation of process parameters is critical to ensuring dimensional accuracy and minimising defects. Various optimisation models, such as Multiple Response Optimization (MRO) techniques combined with Multiple Linear Regression (MLR) models, can effectively manage the interactions between equipment settings and operating conditions in the thermoforming process of polystyrene sheets. These models have demonstrated substantial improvements in process control and product quality by providing precise predictions and adjustments that enhance final product accuracy \cite{Leite2018}. However, while they perform well in the modelling and optimisation of pre-set conditions, they may not fully allow for real-time adjustments. Digital twins, for instance, have been applied to thermoforming processes to model and optimise material consumption, highlighting their potential to enhance sustainability by improving production performance \cite{Turan2022}. Real-time data from sensors and Programmable Logic Controllers (PLCs) were integrated to create a virtual replica of the production process, enabling significant reductions in material waste, scrap ratios, and overall production costs. Nevertheless, the real-time application of digital twins in industrial settings remains limited for reasons including the complexity of integrating these models with existing production systems.

One of the challenges in applying machine learning in the thermoforming process is data scarcity and heterogeneity. Methods investigated to ameliorate these issues include a sim-to-real transfer learning framework utilising a Convolutional Variational Autoencoder (ConvVAE). This method enables accurate predictions of product quality metrics, such as thickness distribution, despite limited and heterogeneous data. By leveraging both structured sensor data and high-dimensional thermal images, this framework is particularly relevant for environments where data variability is a concern \cite{Ramezankhani2024}. However, the reliance on simulated data for training might make application in real-world scenarios challenging. 

The complex and variable conditions inherent in thermoforming processes, such as differential heating and material deformation, underscore the challenge of modelling such processes. Particularly if seeking scalable, robust and generalisable models to ensure consistent product quality across diverse environmental and operational conditions. Artificial neural network (ANN) algorithms have, for example, been used to predict and optimise process parameters in the thermoforming of composite thermoplastics through numerical simulation, focusing on factors such as laminate orientation and tensioner stiffness. These models have been effective in minimising defects and ensuring consistent product quality, significantly reducing the reliance on traditional trial-and-error methods \cite{Tan2022}. However, the substantial computational resources required for these simulations can pose challenges in terms of scalability and applicability across different operational settings.

 Various ML algorithms, including random forest and gradient boosting regressors, have been investigated to predict form accuracy in vacuum-assisted hot-forming processes to produce curved glass for mirrors or head-up displays\cite{Vogel2022}. Different types of input data (set parameters, sensor data, and thermographic images) were used to train ML models. However, the manual feature engineering/extraction may limit the generalisability of these findings to other scenarios. Convolutional neural networks (CNNs) have been used in quality control for thermoformed food packaging, leading to improvements in defect detection accuracy \cite{Banus2021}. However, the potential of CNNs to suggest effective process adjustments for maintaining product quality remains largely unexplored. 


\section{Methodology}\label{sec3}
\subsection{Dataset Collection}\label{subsec31}
In vacuum thermoforming, data collection involves a range of parameters such as heat profiles, vacuum pressure, mould characteristics, and material properties. These detailed parameters are vital for process optimisation and product quality, yet they are seldom shared externally due to industry competitiveness. We, therefore, produced our own dataset. We manufactured samples using a conventional vacuum-forming machine while varying heating power, heating time and vacuum time. Following the forming process, we used a digital camera to capture the visual presentation of the samples. Then, by leveraging machine learning techniques, we use the visual data to enhance the vacuum-forming process, especially in product quality. This approach is tailored to mimic human operator work in doing quality control and improvement.

Vacuum-formed samples were fabricated using a widely available commercial vacuum former machine (Formech 508DT, Formech International Ltd). A hemispherical mould with a radius of 30 mm was additively manufactured using ABS polymer (3D FilaPrint ABS-X 1.75 mm). The training dataset was built based on this simple mould configuration. The forming process was performed manually by heating the sheet materials, vacuum-assisted forming the heated sheet into the mould, cooling it, and demoulding the formed parts. High-impact polystyrene (HIPS) sheets, a common thermoforming material, were employed in this study.

When producing samples, we explored different combinations of forming process parameters, as shown in Table~\ref{tab1}. Subsequently, we recorded those combinations of parameters to create a labelled dataset, which serves as the foundation for training our models. Each combination of parameters produced different quality outcomes in the formed part. The systematic variation of heating power and time is used to determine the formability of the selected materials, as the optimal forming temperature ranges from 120-180 C, depending on the material thickness. Through the selected range of process parameters, we can capture under and overheating scenarios, thereby providing comprehensive data on how different heating levels affect part quality. Meanwhile, vacuum time variation is also critical for achieving desired end shapes. The other affecting factors, such as vacuum pressure and cooling conditions, are set to be constant.

\begin{table}[b]
\caption{Process parameters used for data collection.}\label{tab1}%
\begin{tabular}{@{}llllll@{}}
\toprule
Parameters & Values\footnotemark[1] \\
\midrule
Heating power (\%)  & 40, 50, 60, 70, 80, 100  \\
Heating time (s)  & 10, 15, 17, 20, 25, 30, 40, 50, 60, 75, 90, 105, 120  \\
Vacuum time (s)  & 3, 5, 7 \\
\botrule
\end{tabular}
\footnotetext[1]{The values are the design values, which may differ from the actual experimental value as the forming process is conducted manually.}
\end{table}

\begin{figure}[t]
\centering
\includegraphics[width=1\textwidth]{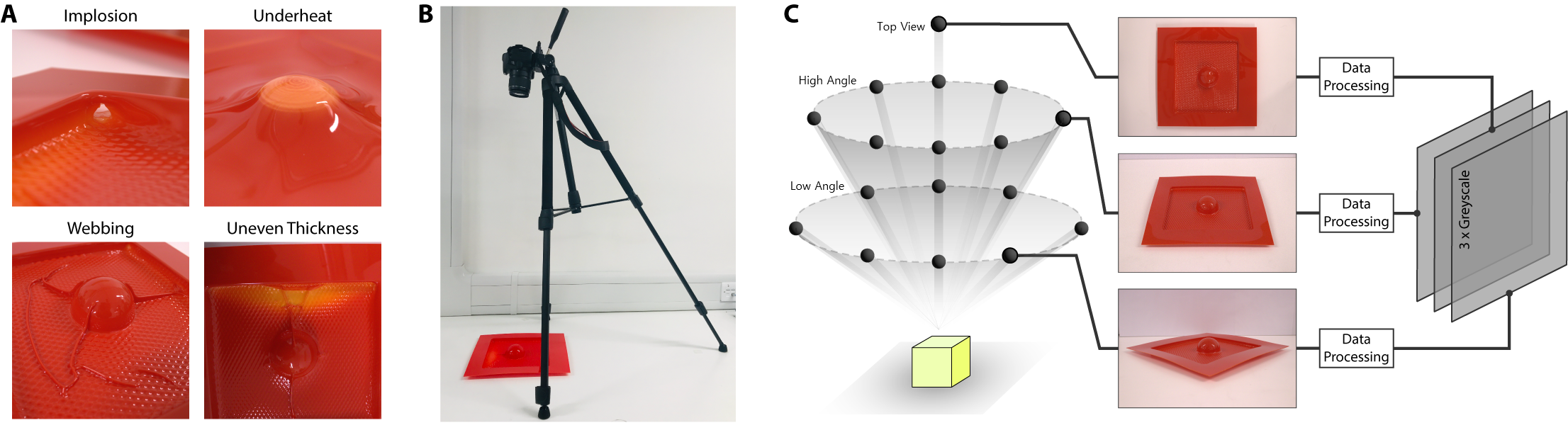}
\caption{Data procedures; A. Failure Modes; B. Data collection setup; C. Schematic diagram for generating training dataset}\label{fig2}
\end{figure}

Comprehensive visual data were collected to assess the quality of the formed parts. Detailed visual data was acquired for each sample by capturing 17 images, consisting of one top view and two sets of angled views (low and high), with each set containing eight images taken at approximately 45-degree increments of sample rotation, as shown in the schematic Fig.~\ref{fig2}-B and Fig.~\ref{fig2}-C. This comprehensive imaging strategy ensured that the machine learning model had access to complete and detailed views of each sample's surface features, including potential defects such as implosion, underheating, webbing, and uneven thickness, as shown in Fig.~\ref{fig2}-A. The images were captured using a Canon EOS 250D camera with an EF-S 18-55mm III macro 0.25m/0.8ft lens under standard room lighting conditions, resulting in a 6024 x 4020 pixels resolution. No specific image positioning setup was employed aside from ensuring that each formed sample was fully captured in the frame with consistent camera positions and settings, particularly regarding zooming or focal length value. This practical and accessible setup ensures that our methods can be easily replicated and applied in various manufacturing settings.

The imaging protocol provided raw data for training the CNNs employed here. CNNs are well-suited for this task due to their ability to automatically learn and extract relevant features from images, making them ideal for quality assessment applications where visual data is abundant. Vision-based machine learning models typically take a three-channel image (RGB values) as input. However, in the case of vacuum forming observations, we do not expect separate RGB values to provide significant information to the model. Instead, the RGB values are merged, and three separate images are input into the model. This allows us to combine information from three different views, increasing the diversity and information-content provided to the model. Combining three of the 17 collected images for each sample will result in 680 unique three-channel images per sample. 

\begin{figure}[t]
\centering
\includegraphics[width=1\textwidth]{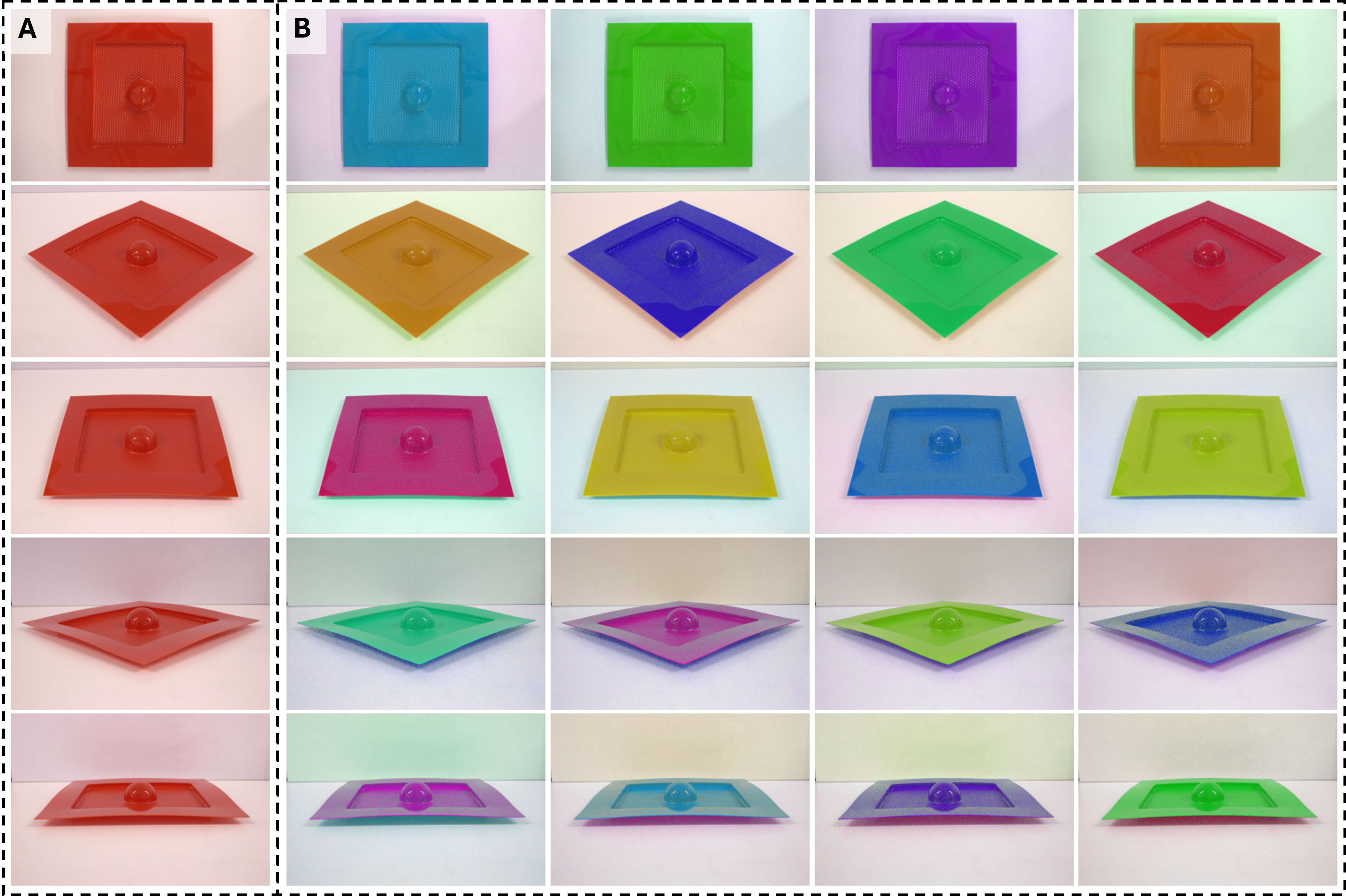}
\caption{Automatic Domain Randomisation (ADR); A. Original sample; B. Samples generated using ADR, where background and sample regions are treated with different augmentations.}
\label{fig3}
\end{figure}

Prior to the combination process, we employed the Automatic Domain Randomisation (ADR) technique to modify image attributes, resulting 
in multiple segmentation with different colours, as shown in Fig~\ref{fig3}. Subsequently, the modified images were converted to greyscale images, reducing the input data's dimensionality into the single channel image without sacrificing essential information like structural features. This technique was applied to all 17 views collected images so it can be further combined into new unique three-channel images. This process was utilised to expose the model to a variety of visual conditions, improving its ability to generalise across different data scenarios, such as different sheet colours or images taken under lighting conditions.

In total, the generated dataset was built from 70 formed-part samples. With the aforementioned data preparation and processing, we obtained 47,600 three-channel images from the unique combinations of original collected images. This extensive dataset was divided into two groups, i.e., the training and testing datasets. The training dataset was built from 65 samples, resulting in 44,200 unique image data, while the rest of the data were used for testing purposes.

\subsection{Training Algorithm}\label{sssec33}
The model training process in this study involved a combination of k-Nearest Neighbour (k-NN) and Convolutional Neural Networks (CNNs). This two-stage approach integrated clustering and regression, aiming to predict the necessary adjustment to the vacuum-forming process parameters to improve the quality of produced parts optimally.

\begin{figure}[t]
\centering
\includegraphics[width=1\textwidth]{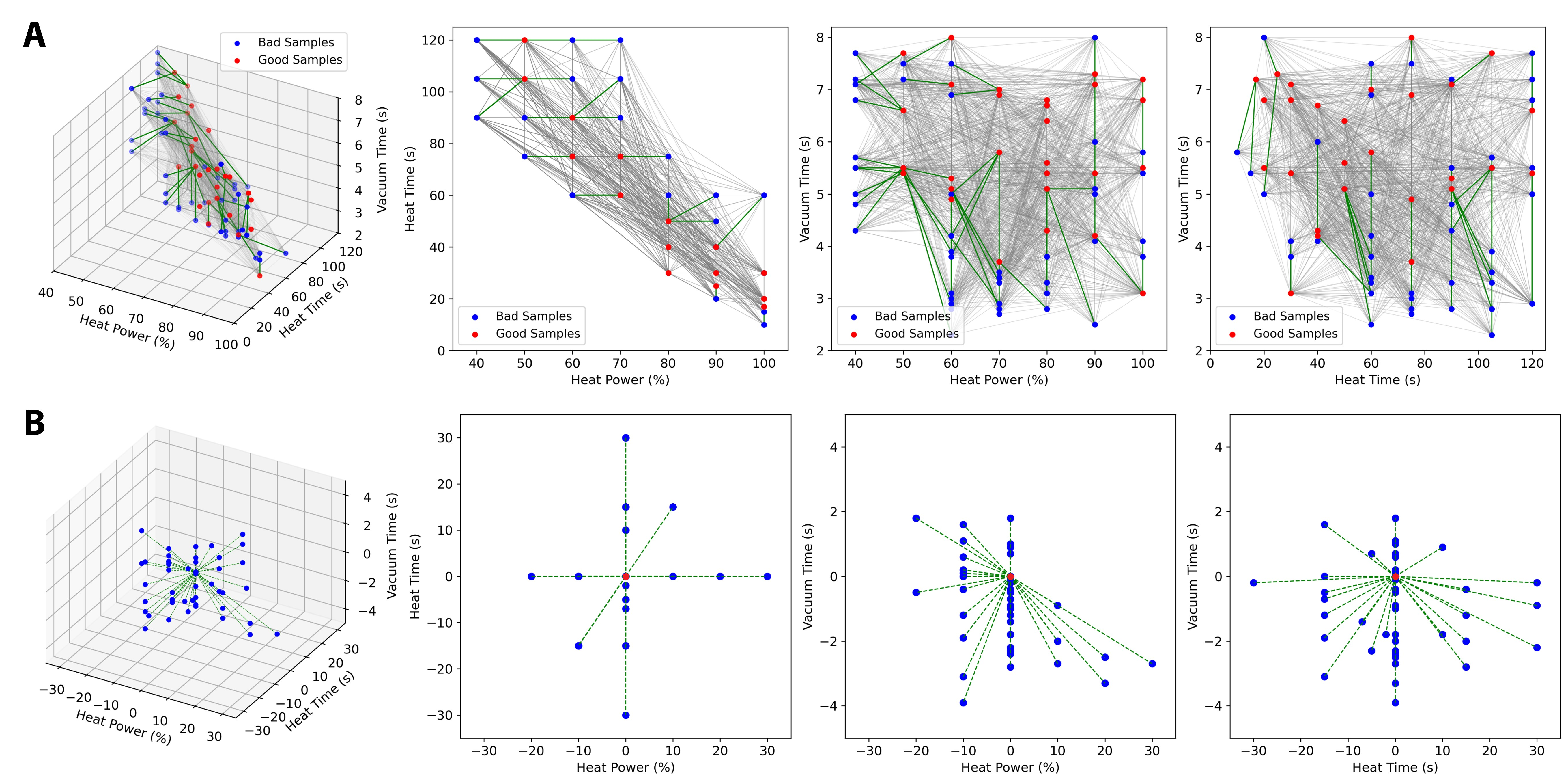}
\caption{k-Nearest Neighbour for Data Labelling. A. Samples in process parameter space. B. Samples in parameter changes space.}\label{fig4}
\end{figure}

In the initial phase, k-NN was employed to establish a correlation between the current set of process parameters and the parameter changes required for a part that was defective to become conforming (good). The process parameters heating power, time, and vacuum time—were recorded for each sample. Samples were categorised as good or bad based on visual appearance and geometric fidelity to the mould as judged by a human expert operator. Fig.~\ref{fig4}-A shows the distribution of clustered good and bad samples. 

To find the necessary adjustments for improving a bad sample, k-NN identifies the nearest good sample in the parameter space. The "nearest" good sample was determined based on a distance metric, Manhattan distance, which measured the similarity between parameter sets. This method is particularly advantageous for recognising local patterns where multiple solutions are feasible, making it highly suitable for this study. The difference in parameters between the bad sample and its nearest good neighbour was calculated and normalised, as plotted in Fig.~\ref{fig4}-B. This normalised difference vector, representing the necessary adjustments (e.g., reductions or increases in heating power, time, or vacuum time), was assigned as the label for the bad sample. Good samples were labelled with the vector (0,0,0), indicating no adjustment is needed as they already meet the quality standards. The decision to use k-NN was based on its simplicity, transparency, and straightforward implementation, making it an accessible and understandable option for guiding similar case studies in the manufacturing process.

The second phase involved training a CNN to predict changes in process parameter vectors. The CNN's objective is to learn the complex relationship between the visual features in the images and the process parameter adjustments represented by the labels. The training dataset consisted of unique combined images of both good and bad samples, with the corresponding adjustment vectors (labels) obtained from the k-NN phase. The CNN architecture, which includes convolutional layers and custom fully connected layers, is designed to learn and extract relevant visual patterns and defects that correlate with the quality issues in the samples. The custom fully connected layers included a regression head comprising three output neurons, each corresponding to a key process parameter: heating power, heating time, and vacuum time. This configuration allowed the CNN to predict specific parameter values based on the input image data.

During training, the CNN processes each image from the generated dataset to predict an adjustment vector. The network is trained using a supervised learning approach, where the predicted vector is compared to the actual label (the adjustment vector from k-NN). The loss function was calculated using Mean Squared Error (MSE), shown in Equation~\ref{eq1}, to measure the difference between the CNN's predicted adjustment vectors and actual labels generated from the k-NN mapping. The model used backpropagation and optimisation techniques to update the weights based on the gradients of the loss function. The gradients are calculated using the chain rule in Equation~\ref{eq2}, allowing the model to adjust its weights to minimise the loss function iteratively. The optimisation algorithm employed the Adam optimiser, an adaptive learning rate optimisation algorithm. Adam combines the benefits of two other extensions of stochastic gradient descent, which maintains a learning rate for each parameter and adapts them based on the mean and variance. This training process enables the CNN to learn the complex mappings from visual image characteristics to the necessary process parameter adjustments.
\begin{equation}
    L=\frac{1}N\sum\limits_{i=1}^{N}(y_i-\hat{y_i})^2\label{eq1}
\end{equation}
\begin{equation}
    \frac{\partial L}{\partial W^{(l)}} = \frac{\partial L}{\partial \hat{y}} \frac{\partial \hat{y}}{\partial a^{(l+1)}} \frac{\partial a^{(l+1)}}{\partial z^{(l+1)}} \frac{\partial z^{(l+1)}}{\partial W^{(l)}}\label{eq2}
\end{equation}
where: $L$ is Loss, $N$ is the number of samples, $y_i$ and $\hat{y_i}$ are true and predicted values, respectively, for the $i$-th sample, and $W^{(l)}$ is weight $W$ in layer $l$.

Apart from the explained process, normalisation techniques were also applied to the input dataset to improve the neural network's training stability and model performance. The input vision data was normalised using a technique adapted from Local Response Normalisation \cite{NIPS2012_Alex}, which standardises pixel values based on the dataset's statistical properties, ensuring consistent input scaling and facilitating improved model convergence. Meanwhile, min-max normalisation was employed within a three-dimensional parameter space for the output space representing the predicted process parameters. This normalisation scales the output predictions to a specified range, which is particularly useful when dealing with outputs with different units or ranges. It ensures that all output parameters contribute equally to the loss calculation and gradient updates, preventing any single parameter from dominating the learning process due to its larger numerical range.

Furthermore, various data augmentation techniques were utilised for the three-channel image dataset input to increase the model's robustness and ability to generalise to diverse data scenarios. These augmentations included per-channel affine transformations, such as scaling, rotation, and translation, along with colour jittering, which involved random adjustments to image brightness, contrast, and saturation. By incorporating these augmentations, the model was exposed to a broad spectrum of input conditions, thereby improving its capacity to handle variations in real-world data.

Overall, the proposed method provides a robust framework for automating and enhancing quality control and process optimisation in vacuum forming. This methodology demonstrated the potential of advanced machine learning techniques in enhancing industrial manufacturing practices, offering substantial efficiency and product quality benefits.

\section{Results and Discussion}\label{sec5}
\begin{figure}[t]
\centering
\includegraphics[width=1\textwidth]{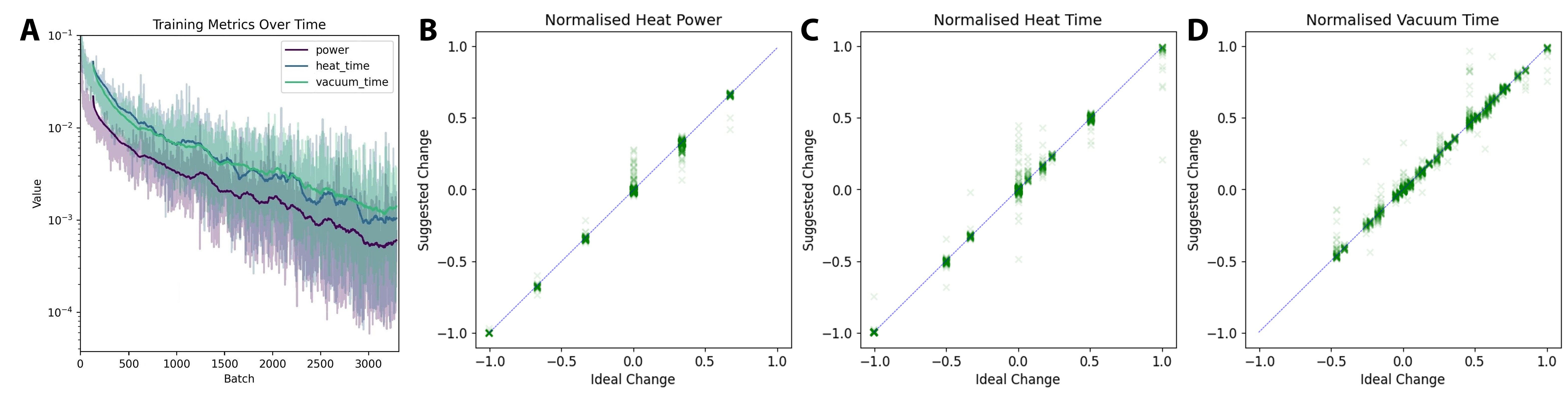}
\caption{Training metrics; A. Loss evolution during training; In B, C, and D, we show predicted vs ideal changes for heat power, heat time and vacuum time, respectively. }\label{fig5}
\end{figure}

Here, we describe model training, validation, and testing to optimise the vacuum forming process. The discussion is separated into three areas: training performance, validation metrics, and performance testing across different conditions, including domain-specific and cross-domain datasets.
Fig.~\ref{fig5}-A illustrates the training metrics over time, highlighting the loss convergence across different process parameters, including heating power, heating time, and vacuum time. The model underwent training for approximately 3000 batches, where the primary objective was to minimise the mean squared error (MSE) between predicted and actual parameter adjustments.

The training curves for all three parameters (heat power, heat time, and vacuum time) showed a consistent decrease in error values as the number of batches increased, indicating effective learning and convergence. Initially, the model experienced a steep decline in error, particularly in the first 1000 batches, which is typical as the model learns the most significant features early on. Beyond this point, the error reduction rate slowed, signalling the model's progression towards fine-tuning the parameter space.
The curves also exhibited slight fluctuations, particularly in the vacuum time metric, suggesting some instability. 

The training process of the proposed vision-based machine learning model demonstrated robust stability and effective convergence. Observations from the training and validation loss curves indicated a consistent decrease in error rates without significant fluctuations throughout the training epochs. Minor variations observed in individual training batches were attributed to the diversity of data within each batch and did not indicate any underlying instability in the learning process.

The model achieved convergence after approximately 13 epochs, as evidenced by the stabilisation of the validation loss. An early stopping mechanism was employed, halting the training at around 2,500 batches once the validation loss plateaued. This approach effectively prevented overfitting, ensuring that the model maintained generalisability when exposed to unseen data. However, the training loss continued to decrease beyond this point; hence, early stopping prioritised the model's performance on validation data over minimal improvements on the training set.

To further enhance the model's learning efficiency and prevent overfitting, weight decay regularisation and a learning rate scheduler were integrated into the training process. Weight decay regularisation penalised large weights in the network, promoting simpler models that generalise better to new data. The learning rate scheduler dynamically adjusted the learning rate during training, enabling the model to make larger updates initially and finer adjustments as it approached convergence. These techniques collectively contributed to the model's stable and efficient training performance.

\subsection{Performance Test}\label{subsec52}
The performance of the vision-based machine learning model in predicting optimal process parameter adjustments for vacuum forming was evaluated using a diverse testing dataset. This dataset differed significantly from the training dataset, which only included red sheets with a thickness of 1 mm formed over a hemispherical mould. The testing dataset introduced additional challenges by incorporating three different mould shapes (a cut cone and two tapered boxes of different heights) and various sheet colours and thicknesses, including purple (1 mm), green (1.5 mm), and orange (2 mm) sheets.

\begin{figure}[t]
\centering
\includegraphics[width=1\textwidth]{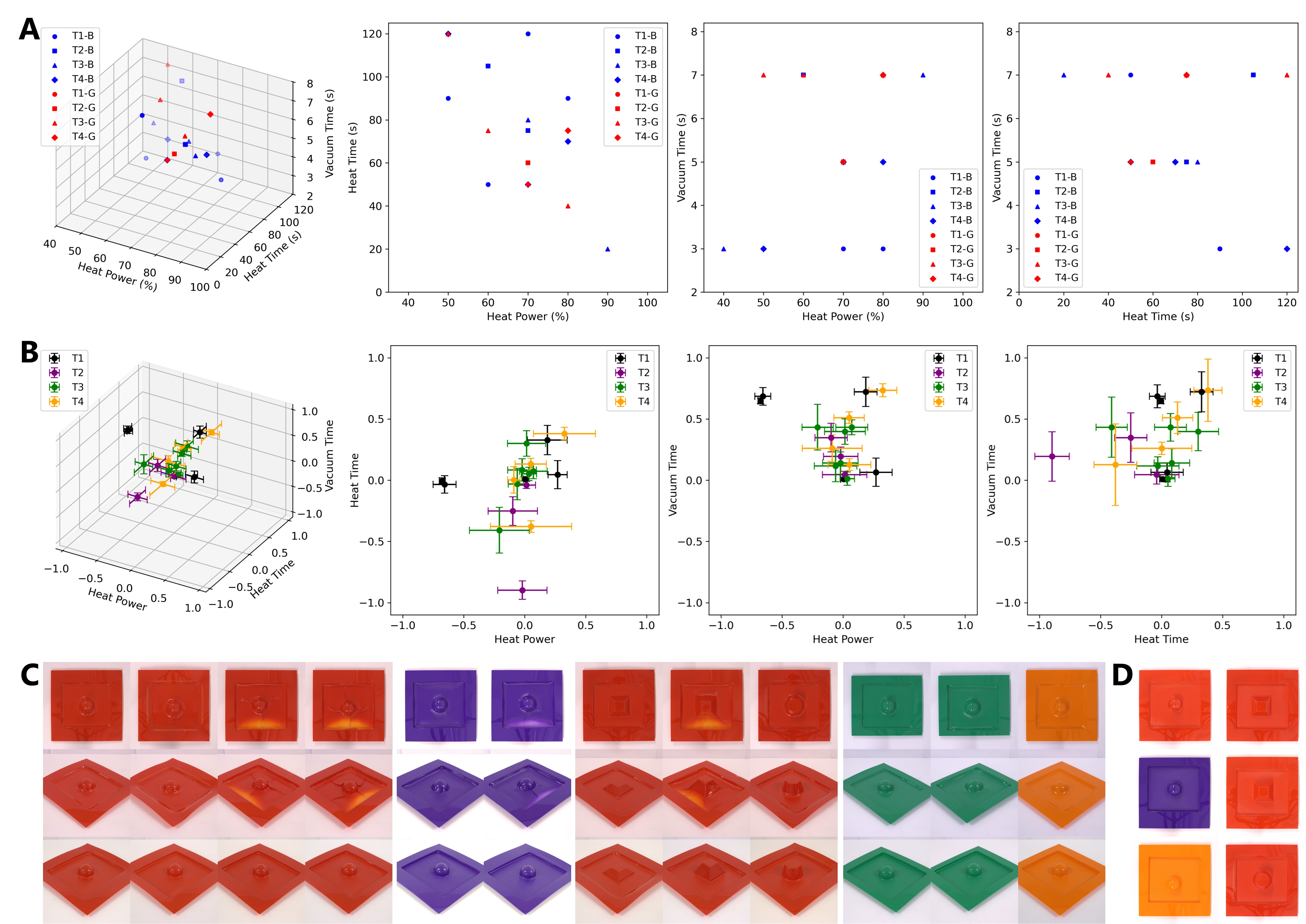}
\caption{Testing results; A. Process parameters map for a testing dataset; B. Normalised suggested process parameters changes; C. Qualitative judgment of the testing results showing formed parts with initial process parameters (first and second row) and with new process parameters based on the suggested changes (last row); D. Formed parts detected as good results}\label{fig6}
\end{figure}

As illustrated in Fig.~\ref{fig6}-A, the test domain parameters covered a wide range of heat power, heat time, and vacuum time variability. This variability ensured that the model's predictions were tested across a broad spectrum of scenarios, reflecting real-world manufacturing conditions. The diversity in mould shapes, colours, and sheet thicknesses posed a significant challenge, as these factors directly influence the formability of the material and the accuracy of the vacuum-forming process.

Fig.~\ref{fig6}-B displays the model's suggested changes in process parameters for each test case. The CNN-trained model was tasked with predicting adjustments in heating power, heating time, and vacuum time. These predictions were aimed at improving the quality of the vacuum-formed parts, as visualised in the images in Fig.~\ref{fig6}-C. The transition from the second row (representing poor quality) to the third row (representing improved quality) in these images demonstrates the model's effectiveness. Despite the complexity of the different test cases, the model consistently suggested parameter changes that led to noticeable quality improvements.

However, some discrepancies were observed in the model's predictions, particularly when dealing with different sheet thicknesses. For instance, the suggested parameters did not always produce perfect samples when forming thicker sheets. This indicates that while the model performs well within the tested domain, further model refinement, training data, or possibly the addition of a heuristic scaling factor would be necessary to predict the required adjustments for materials with different thicknesses accurately.

By predicting and adjusting process parameters in real-time, the model, or a model adopting this methodology, could help to minimise scrap rates and improve production efficiency. However, as highlighted, the model currently demonstrates good generalisation across different colours and standard shapes but would likely require further development to handle more complex geometries and materials of varying thicknesses and thermal conductivities effectively. The final set of results, presented in Fig.~\ref{fig6}-D, depicts the good parameters dataset, which did not require any adjustments. This supports the model's ability to identify when no change is needed, ensuring that process stability is maintained where appropriate.

The model's performance was extensively evaluated across various test domains to assess its scalability and adaptability. When tested on datasets involving different geometries and colours but maintaining the same material thickness as the training data, the model consistently provided accurate and valuable suggestions for process parameter adjustments. This suggests that the model effectively learned the underlying relationships between visual features and optimal process parameters within the trained domain.

The model exhibited limitations when applied to scenarios involving different material thicknesses not represented in the training dataset. Predictions for samples with varying thicknesses lacked the accuracy observed in same-thickness scenarios, underscoring the model's sensitivity to parameters outside its trained experience. This performance discrepancy is primarily attributed to the homogeneity of the training data in terms of material thickness. Incorporating a more diverse training dataset encompassing various thicknesses and material properties would likely enhance the model's generalisability and predictive accuracy across a broader range of manufacturing conditions.

Despite these limitations, the model's suggestions in unfamiliar domains still provided valuable insights that could assist in approximating suitable manufacturing parameters. Future work should focus on expanding the training dataset's diversity and incorporating continual learning approaches to progressively refine the model's adaptability and performance. Data augmentation techniques played a crucial role in enhancing the model's robustness and ability to generalise across different visual conditions. The application of Automatic Domain Randomization (ADR) introduced controlled variations in the training images, such as changes in lighting conditions and colour schemes. This exposure to a broader range of visual scenarios enabled the model to maintain high-performance levels even when confronted with test data exhibiting slight environmental or appearance differences from the training set.

The effectiveness of ADR was particularly evident in the model's consistent performance across different colours of the same material and geometry. By simulating various conditions during training, ADR prepared the model to handle real-world variability, reducing its dependence on strictly controlled imaging environments. While current data augmentation strategies successfully addressed minor environmental variations, future research could explore more extensive augmentation techniques to bolster the model's resilience further. This includes introducing more significant variations in viewing angles, backgrounds, and possibly simulating defects or imperfections to prepare the model for a wider array of practical scenarios.

The rapid prediction capabilities of the model, generating suggestions within seconds, highlight its potential applicability in real-time manufacturing environments. Integrating such a system could facilitate swift adjustments to process parameters, thereby enhancing production efficiency and product quality. However, current limitations in data acquisition processes, particularly the requirement for prescribed image views under controlled conditions, pose challenges for seamless real-world implementation.

Addressing these challenges will involve developing more flexible and automated data collection methods capable of capturing relevant visual information under varied and less controlled conditions. Incorporating multiple cameras and sensors to gather diverse perspectives and environmental data could enhance the system's practicality and responsiveness in dynamic manufacturing settings. Future developments should also consider transitioning from normalised, non-dimensional parameter suggestions to absolute units. Providing recommendations in standard units would improve the intuitiveness and direct applicability of the model's outputs, facilitating easier integration with existing manufacturing control systems.



\section{Conclusion}\label{sec7}
The developed vision-based machine learning model demonstrates promising capabilities in optimising vacuum forming processes through effective and efficient prediction of process parameter adjustments. While showing robust performance within its trained domains, particularly across various geometries and colours, the model's limitations in handling untrained material thicknesses underscore the need for more diverse training data and adaptive learning strategies. Data augmentation techniques like ADR have proven beneficial in enhancing model robustness, and future work should focus on refining data acquisition methods, integrating physically meaningful parameter outputs, and implementing continual learning frameworks to fully realise the system's potential in real-world manufacturing environments. These advancements will contribute significantly to the progression towards intelligent and autonomous manufacturing systems envisioned in Industry 4.0.

\backmatter


\bmhead{Acknowledgements}
The authors acknowledge financial support from the Made Smarter Innovation - Research Centre for Connected Factories (EP/V062123/1). AK is supported by the Jardine Scholarship from the Jardine Foundation.

\section*{Declarations}

\subsection*{Funding}
This work was supported by the Made Smarter Innovation - Research Centre for Connected Factories (EP/V062123/1).

\subsection*{Conflict of Interest}
SWP is a co-founder of Matta Labs, a spin-out company developing AI-based software for manufacturing. Other authors declare that they have no known conflict of interest or personal relationships that could have appeared to influence the work reported in this paper.

\subsection*{Data Availability}
The datasets generated during the current study are not publicly available due to self-built datasets but are available upon a reasonable request. For review purposes, the vacuum forming image dataset sample can be accessed at this link: \url{https://bit.ly/sample_VFD}.

\subsection*{Author Contribution}
All authors have contributed to the conceptualisation, design and implementation of the research activities. Andi Kuswoyo specifically carried out the experimental parts, including fabricating the sample to collect the dataset. He also wrote the manuscript with the support of all other authors. Christos Margadji contributed mostly to the ML model training. Sebastian Pattinson helped to supervise the project.





\bibliography{sn-bibliography}

\end{document}